\documentclass[sigconf]{acmart}

\usepackage{booktabs} 
\usepackage{hyperref}
\usepackage{url}
\usepackage{array}
\usepackage{color}

\copyrightyear{2018}
\acmYear{2018}
\setcopyright{acmcopyright}
\acmConference[MM '18]{2018 ACM Multimedia Conference}{October 22--26, 2018}{Seoul, Republic of Korea}
\acmPrice{15.00}
\acmDOI{10.1145/3240508.3240635}
\acmISBN{978-1-4503-5665-7/18/10}

\begin{document}
\title{USAR: an Interactive User-specific Aesthetic Ranking Framework for Images}

\author{Pei Lv$^*$}
\affiliation{%
  \institution{Zhengzhou University}
  \streetaddress{100 Kexue Road}
  \city{Zhengzhou}
  \state{Henan}
  \postcode{450000}
}
\email{ielvpei@zzu.edu.cn}

\author{Meng Wang$^*$}
\affiliation{%
  \institution{Zhengzhou University}
  \streetaddress{100 Kexue Road}
  \city{Zhengzhou}
  \state{Henan}
  \postcode{450000}
}
\email{vergilwm@gs.zzu.edu.cn}

\author{Yongbo Xu}
\affiliation{%
  \institution{Zhengzhou University}
  \streetaddress{100 Kexue Road}
  \city{Zhengzhou}
  \state{Henan}
  \postcode{450000}
}
\email{risingdragon@live.cn}

\author{Ze Peng}
\affiliation{%
  \institution{Zhengzhou University}
  \streetaddress{100 Kexue Road}
  \city{Zhengzhou}
  \state{Henan}
  \postcode{450000}
}
\email{zpeng@ha.edu.cn}

\author{Junyi Sun}
\affiliation{%
  \institution{Zhengzhou University}
  \streetaddress{100 Kexue Road}
  \city{Zhengzhou}
  \state{Henan}
  \postcode{450000}
}
\email{jysun@ha.edu.cn}

\author{Shimei Su}
\affiliation{%
  \institution{Zhengzhou University}
  \streetaddress{100 Kexue Road}
  \city{Zhengzhou}
  \state{Henan}
  \postcode{450000}
}
\email{smsu@zzu.edu.cn}

\author{Bing Zhou}
\affiliation{%
  \institution{Zhengzhou University}
  \streetaddress{100 Kexue Road}
  \city{Zhengzhou}
  \state{Henan}
  \postcode{450000}
}
\email{iebzhou@zzu.edu.cn}

\author{Mingliang Xu}
\affiliation{%
  \institution{Zhengzhou University}
  \streetaddress{100 Kexue Road}
  \city{Zhengzhou}
  \state{Henan}
  \postcode{450000}
}
\email{iexumingliang@zzu.edu.cn}

\begin{abstract}

When assessing whether an image is of high or low quality, it is indispensable to take personal preference into account. Existing aesthetic models lay emphasis on hand-crafted features or deep features commonly shared by high quality images, but with limited or no consideration for personal preference and user interaction. To that end, we propose a novel and user-friendly aesthetic ranking framework via powerful deep neural network and a small amount of user interaction, which can automatically estimate and rank the aesthetic characteristics of images in accordance with users' preference. Our framework takes as input a series of photos that users prefer, and produces as output a reliable, user-specific aesthetic ranking model matching with users' preference. Considering the subjectivity of personal preference and the uncertainty of user's single selection, a unique and exclusive dataset will be constructed interactively to describe the preference of one individual by retrieving the most similar images with regard to those specified by users. Based on this unique user-specific dataset and sufficient well-designed aesthetic attributes, a customized aesthetic distribution model can be learned, which concatenates both personalized preference and aesthetic rules. We conduct extensive experiments and user studies on two large-scale public datasets, and demonstrate that our framework outperforms those work based on conventional aesthetic assessment or ranking model.

\end{abstract}



\maketitle

\section{Introduction}

The proliferation of social networks and mobile devices with cameras has led to an explosive increase in the number of digital images. This has generated large personal photograph datasets for maintaining beautiful memories. However, the organization of an ideal personal album or collection manually from such massive number of images is onerous, and this task is always time-consuming and challenging. The major problem underlying this challenge is the accurate recognition of the personal aesthetic preference of different users. In this paper, we study how to automatically assess the aesthetic characteristics of images through taking into account the user's preference in a simple interactive way.

\begin{figure}[!t] \begin{centering}
  \centering
  \includegraphics[scale=.30]{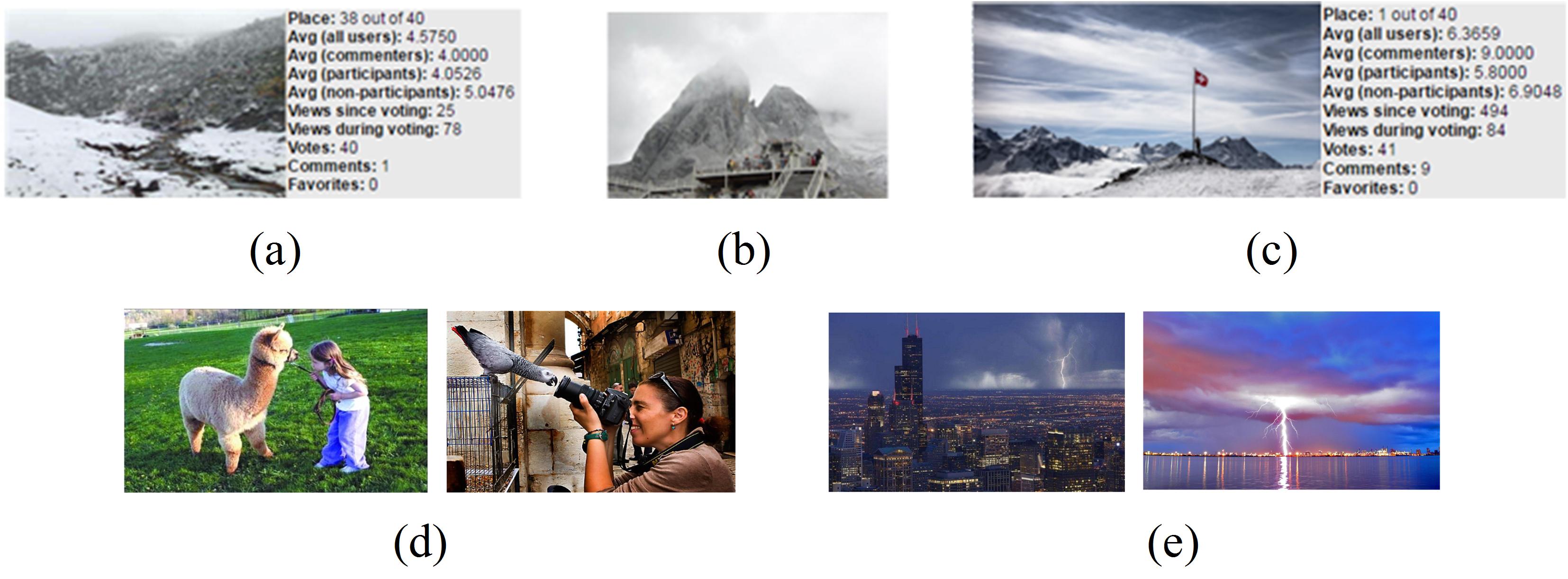}
  \centering
  \caption{(Top row) The constraints of conventional aesthetical binary classification. (a) is deemed as low quality while (c) scores much higher. It is difficult to quantify the actual quality of (b). (Bottom row) Personal preference varies in different individuals. One of our attendants prefers images of human portraits in (d) while another one likes images of lightning in (e)}
  \label{hand_crafted}
  \end{centering}
\end{figure}

In recent years, plenty of methods have been proposed to measure the aesthetic quality of photographs. Most researchers focus their attention on selecting and setting universal descriptors derived from high quality images. Based on the assumption that high quality images share certain common aesthetic rules, massive hand-crafted low level visual features~\cite{8,3,4,9,5,17,6,10,7} and high level aesthetic attributes~\cite{7,8,9,10,11,21}  have been proposed. However, traditional hand-crafted visual features are limited and restrictive due to the following reasons: 1) These man-made visual features are the approximations of aesthetic rules, failing to fully capture the aesthetic abstract. 2) The universal visual features lack of consideration for subjectivity and personal preference of different users. For example, as shown in the first row of Fig.~\ref{hand_crafted}, Fig.~\ref{hand_crafted}(a) is regarded as low quality while Fig.~\ref{hand_crafted}(c) always receives more attention. However, it is difficult and ambiguous to decide whether Fig.~\ref{hand_crafted}(b) is of low quality or of high quality. For the second row, when asked to choose their favorite image, some of our respondents preferred Fig.~\ref{hand_crafted}(e) to Fig.~\ref{hand_crafted}(d), while other users reported the opposite choice. Since aesthetics is highly subjective and complex, each user has his or her own judgment of what is beautiful.

Recently, instead of using traditional hand-crafted visual features, the state-of-art feature extraction technique based on deep learning has been involved to evaluate the aesthetic quality of images~\cite{17,23,33,14}. Compared with those traditional methods, the most notable difference is that the deep features of the input images could be extracted automatically without making any artificial approximations of the aesthetic rules. However, most of above work pay their attention on the task of  binary classification without considering the subjectivity and personal preference of diverse individuals. In ~\cite{32}, Ren et al. try to address this personalized aesthetics problem by showing that individual's aesthetic preferences exhibit strong correlations with image content and aesthetic attributes, and the deviation of individual's perception from generic image aesthetics is predictable. They propose a new approach to personalized aesthetics learning that can be trained even with a small set of annotated images from one user. However, since one user's preference is highly subjective and his/her choice one time is occasional, a small set of annotated images is insufficient to fully represent his or her personal preference.

To solve the above problems, in this paper, we propose a novel and interactive user-friendly aesthetic ranking framework, called User-specific Aesthetic Ranking (USAR), which consists of three modules: primary personalized ranking (PPR), interaction stage (IS) and user-specific aesthetic distribution (USAD). The proposed framework takes as input a series of photos that users prefer, and produces as output a reliable, user-specific aesthetic distribution matching with user's preference. In the module of PPR, a unique and exclusive dataset will be constructed interactively to describe the preference of one individual by retrieving the most similar images with regard to those specified by users. This is based on the fact that the aesthetic preference of one user will remain unchanged for a long time~\cite{31,32}. The powerful Deep Convolutional Neural Network is involved and optimized to retrieve those content similar images through several amounts of interactions in the module of IS. Based on this unique user-specific dataset and sufficient well-designed aesthetic attributes, a customized aesthetic distribution model will be learned in the module of USAD, which concatenates both personalized preference and photography rules. Given an input image, its corresponding aesthetic distribution will be computed by USAD. After that, the correlation coefficient between one user's specific aesthetic distribution and that of input image can be obtained. The larger the coefficient is, the higher aesthetic score and ranking is. We conduct extensive experiments and user studies on two large-scale public datasets, and demonstrate that our framework outperforms those work based on conventional aesthetic assessment or ranking model.

The contributions of this paper mainly focus on the following aspects.

\begin{itemize}
\item We propose a novel and user-friendly aesthetic ranking framework via powerful deep neural network and a small amount of interaction, which can automatically rank the aesthetic quality of images in accordance with user's preference.

\item We propose an efficient and limited interactive method to construct a unique and exclusive dataset to represent the aesthetic preference of one individual, which can overcome the problem of user's subjective preference and occasional choice.

\item We propose a customized aesthetic distribution model based on a unique user-specific dataset and sufficient well-designed aesthetic attributes, which concatenates both user's personalized preference and aesthetic rules.
\end{itemize}

\begin{figure*}[!t] \begin{centering}
  \centering
  \includegraphics[scale=.29]{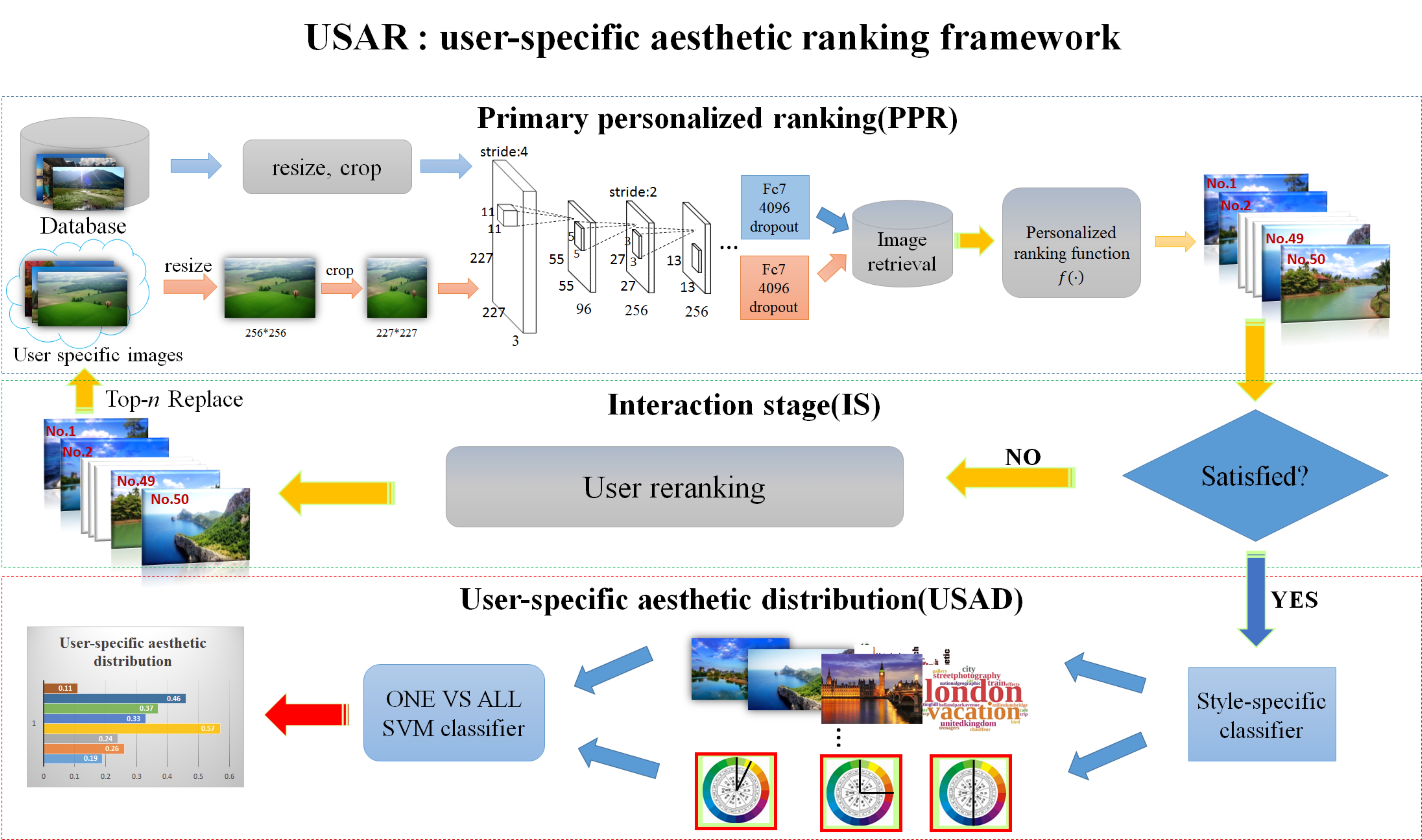}
  \centering
  \caption{The overview of user-specific aesthetic ranking framework. The framework consists of three modules: Primary personalized ranking(PPR), Interaction stage(IS) and User-specific aesthetic distribution(USAD). The PPR is applied to construct a unique and exclusive dataset to represent the aesthetic preference of one individual. The IS is deployed to refine the AlexNet as well as the PPR. Finally, the images ranked by PPR is sent to USAD to generate the user-specific aesthetic distribution.
  }
  \label{algo_overview}
  \end{centering}
\end{figure*}

\section{Related work}

User-specific aesthetic quality ranking mainly concerns two important problems: how to evaluate the aesthetic quality of images and how to collect user-specific images. In this section, we first review related aesthetic quality assessment work, then discuss the problem concerning personalized image searching and ranking, which is used to collect user-specific images.

The previous work mainly focus on hand-crafted visual features that all high-quality images may share. Extensive experiments have been conducted with low-level visual features~\cite{3,4,6,22} and high-level features combinations~\cite{5,7,8,11,12,15,16,18,19,21}. Datta et al.~\cite{3}  utilized a computational approach to understand images using a  56-dimensional features vector. Luo et al.~\cite{4} first proposed subject region extraction on the assumption that professional photographers focus on their subjects while blurring the background. Mavridaki et al.~\cite{6} proposed aesthetic feature pattern, coupled with other features such as simplicity, sharpness, composition and so on. Aydin et al.~\cite{22} presented five kinds of feature (sharpness, colorfulness, tone, clarity, depth) to measure the image quality to enable automatic analysis and editing. Some researchers were not satisfied with the results of low-level visual features, so they have turned their attention to the combination of high-level features. Dhar et al.~\cite{7} proposed high-level describable attributes based on content, composition and sky illumination to predict the interestingness of the input images. Lo et al.~\cite{8} proposed a set of features that are discriminative without adopting any computationally intensive techniques. Luo et al.~\cite{15} extracted different features for different categories of photos and then generated category-specific classifiers. Although hand-crafted features play a certain role in assessing the image quality, it is a man-made approximation of the abstract aesthetic rules and may fail to capture the full diversity and beauty of a photographic image.

In recent years, with the rise of deep learning, the extraction of deep features has gradually become popular for the task of image quality assessment. Lu et al.~\cite{14} developed a double-columned deep convolutional neural network to capture both global and local characteristics of images. Dong et al.~\cite{15} directly utilized a model trained on ImageNet~\cite{17}, acquiring a 4096 dimensions feature vector extracted from the former and  achieved a better performance. Tian et al.~\cite{23} focused on the similarity of images in the same category to propose a query dependent model consisting of both deep visual features and semantic features. Despite the high accuracy of the binary classification tasks that deep neural network achieves, it can not extract the meaning of aesthetics in the absolute sense and understand personal preference. Ren et al. ~\cite{32} tried to address this personalized aesthetics problem by showing that individual's aesthetic preferences exhibit strong correlations with content and aesthetic attributes, and the deviation of individual's perception from generic image aesthetics is predictable. They proposed a new approach to personalized aesthetics learning that can be trained even with a small set of annotated images from a user, but the ability to characterize user's preference is limited and unstable.

Our proposed User-specific Aesthetic Ranking (USAR) framework tries to address above problems. To overcome the shortcomings of low or high level visual features, a powerful and iterative optimized AlexNet is deployed to capture the full diversity of user selected images. Aiming at the problem of unstable personalized ranking of images, we try to construct a unique and exclusive dataset with user's interactions. Based on the unique user-specific dataset and sufficient well-designed aesthetic attributes, a customized aesthetic distribution model is learned, which concatenates both personalized preference and photography rules.

\section{User-specific aesthetic ranking framework}

In this paper, we propose a user-specific aesthetic ranking framework by using a massive image dataset via AlexNet (illustrated in Fig.~\ref{algo_overview}), which consists of three modules: 1) Primary personalized ranking, 2) Interaction stage and 3) User-specific aesthetic ranking. Given a set of preferred images, we first extract their content features for further retrieval of similar images from the whole aesthetic database and construct a retrieval set. Then a primary personalized ranking ${R_{PPR}}$ is generated from the primary personalized ranking module. In order to overcome the instability suffered from the direct use of a small amount of samples, i.e., the user-specific images, we perform refining strategy by asking user to interact with the primary ranking images and treat them as the ground-truth, which are subsequently sent to our style-specific classifier to generate a user-specific aesthetic distribution ${D_{USAR}}$. During the testing stage, the learned ranking module outputs the testing distribution ${D_{test}}$. Consequently, user-specific aesthetic ranking is obtained by calculating the correlation coefficients between ${D_{USAR}}$ and ${D_{test}}$.

\subsection{Primary Personalized ranking}
\label{ppr_sec}
Due to the strong representability of the deep feature, we deploy Deep Convolutional Neural Network to understand images. The features we extracted are based on AlexNet used on ImageNet. It is proposed and designed by Krizhevsky et al.~\cite{17}. As for the extraction on feature of aesthetic attributes, there is no need to construct too deep network. AlexNet is well-matched for the requirement on extracting aesthetic feature as well as computing time while we refine the whole network to generate user-specific aesthetic model. As shown in Fig.~\ref{algo_overview}, we first ask user to pick $m$ images that they prefer, denoted by user-specific images. Then the image retrieval method described below is applied to learn a customized ranking function outputting primary personalized ranking. We adopt the interaction strategy by requiring user to rerank the top $k$ images and replace the original user-specific image with them. This aims at refining the whole AlexNet by reordering user's ranking to fit the personalized choice. The work flow detailed above will repeat $N$ times in an attempt to get the ideal fitting results.

The image retrieval method adopted is detailed as follow. We suppose $U = \{ {I_{u1}},{I_{u2}},..., {I_{ui}}, ..., {I_{un}}\}$ be the set of preferred images that the user chooses, where ${I_{ui}}$ denotes the $i$\textsl{-th} favorite image of the user and $U \in \Gamma$. The $\Gamma  = \{ {I_i}\}, i = 1,2,3...n$ is the entire aesthetic image database. For one given image ${I_{ui}}$, we first extract its visual features for the retrieval of the similar images from $\Gamma$. For the retrieval purpose, we explore the neighbors of the user-specific image in a joint visual space by adopting the equation below:

\begin{equation}
{S^{({I_i})}} = \{ {I_i},{I_i} \in \Gamma  \cap {I_i} \in \Psi \}
\end{equation} where $\Psi$ is neighboring joint visual space of user-specific image. Namely, $\Psi$ is the image dataset containing similar images to the user-specific one.
For each user-specific image, we perform different strategies, dynamically to select ${m}$ similar images that are retrieved from $\Gamma$ and concatenate the images into retrieval result ${S^{({I_i})}}$. Once the retrieval result is obtained, a user-specific ranking function $f(\cdot)$ is learned to predict the level of ${I_u}$.

After the personalized retrieval result ${S^{({I_i})}}$ is generated, the objective of our USAR model is to predict user's preference. Given the aesthetic image database $\Gamma  = \{ {I_i}\}, i = 1,2,3...n$, whose image $I_i$ is represented by the feature $\{ {x_i}\}  = \{ {x_1},{x_2},...,{x_l}\}$, where $l$ denotes the dimension of the features extracted, the personalized ranking function of the USAR model is shown as follow:

\begin{equation}
r({x_i}) = {w^T}{x_i}
\end{equation}

And it can be  maximized to satisfied the following constraints :

\begin{equation}
\forall (i,j) \in \Gamma :{w^T}{x_i} > {w^T}{x_j}
\end{equation}

This approach is similar to that used in SVM classification, where the goal is to generate a relative image pair in accordance with the query identification within. This leads to an optimization problem shown as below:

\begin{equation}
\begin{array}{l}
\min {\rm{ }}(\frac{1}{2}||{w^T}||_2^2 + C(\sum\limits_{}^{} {{\xi _{ij}}} ))\\
s.t.{\rm{   }}{w^T}{x_i} - {w^T}{x_j} \ge 1 - {\xi _{ij}};\forall (i,j) \in \Gamma \\
{\rm{        }}{\xi _{ij}} \ge 0
\end{array}
\end{equation} where $w$ is weight vector on ranking function, $C$ is tradeoff between training error and margin. ${\xi _{ij}}$ is slack variable of different image pairs. We solve this problem by training a linear kernel with ${\rm{SV}}{{\rm{M}}^{rank}}$~\cite{27}. We denote the primary personalized ranking released by the ranking function as  ${R_{PPR}}$.

\begin{figure}[!t] \begin{centering}
  \centering
  \includegraphics[scale=.45]{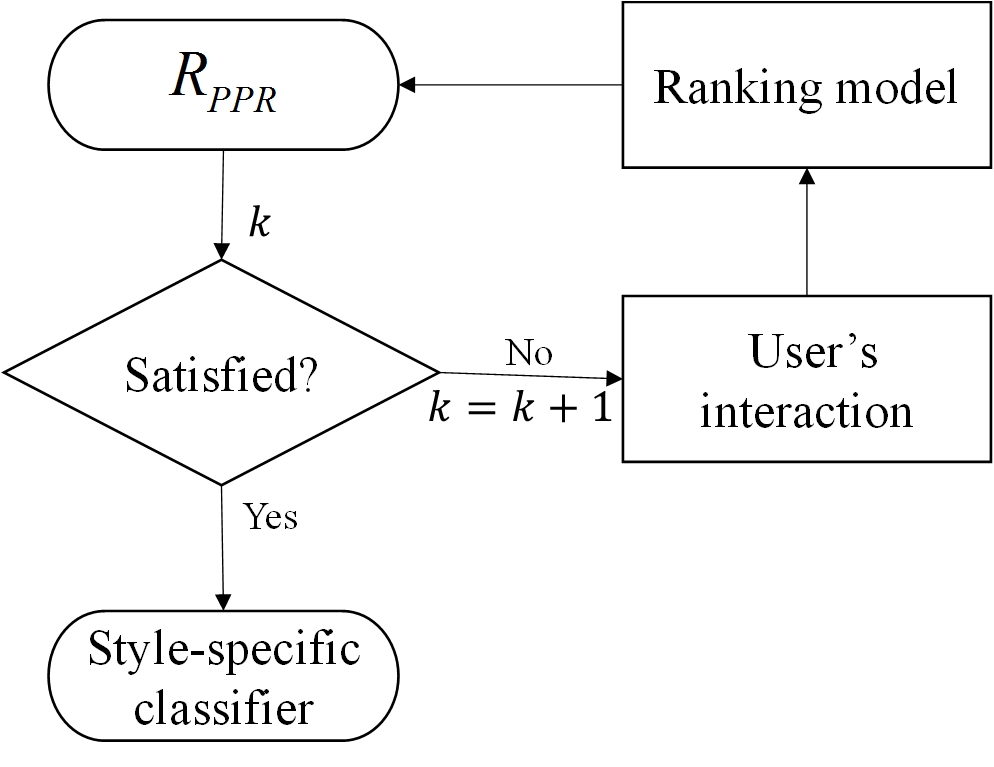}
  \centering
  \caption{The pipeline of the interaction stage. After acquiring the ${R_{PPR}}$, interaction stage is applied to refine the ranking model with user's interaction. If users are satisfied with the ${R_{PPR}}$, we then send the images in ${R_{PPR}}$ to style-specific classifiers. Otherwise, users are allowed to implement a series of operation such as reranking or deleting. The top-$k$ of the new ${R_{PPR}}$ is involved to replace the user-specific images for the refinement of the PPR with $N$ times interaction.}
  \label{interaction}
  \end{centering}
\end{figure}

\subsection{Interaction stage}
\label{is_sec}

Considering the probable instability suffered from small sample directly use of user-specific image, we propose refinement strategy by adding user's interaction on ${R_{PPR}}$. We first ask user to interact with our system by reranking ${R_{PPR}}$  \textsl{N} times and treat the new ${R_{PPR}}$ denoted by ${R_{UPPR}}$ as the prior ground truth data of the user. The concrete workflow shown in Fig.~\ref{interaction} is as follow:

Specifically, after generating the ${R_{PPR}}$, users are required to rerank the ${R_{PPR}}$. We dynamically adjust the reranking times $N$ in accordance with user's interaction. We set $N$ to 0 if users are satisfied with the ${R_{PPR}}$. Otherwise, the reranked ${R_{PPR}}$ is involved to \textsl{Primary personalized ranking generation} stage to regenerate the new ${R_{PPR}}$ denoted by ${R_{UPPR}}$ and $N$ is updated accordingly. Since our goal is to accurately explore and localize the user's preference, we test the performance of ${R_{UPPR}}$ by selecting different ${N}$. Finally, the retuned primary personalized ranking is delivered to \textsl{User-specific aesthetic distribution generation} stage.

\begin{table}[]
\caption{The chosen aesthetic attributes}
\label{aesthetic_tab}
\begin{tabular}{|c|c|c|c|}
\hline
Aesthetic attribute       & Method   & Aesthetic attribute           & Method   \\ \hline
Rule of Thirds            &  \,~\cite{6}  & Tone                          &  \,~\cite{7}  \\ \hline
Center composition        &   \,~\cite{6}  & Use of light                  & \,~\cite{15}\\ \hline
HROT &   \,~\cite{6}  & Saturation                    &  \, ~\cite{3}  \\ \hline
Sharpness                 &   \,~\cite{6}  & Image size                    &  \, ~\cite{35} \\ \hline
Pattern                   &   \,~\cite{6}  & Edge Composition              &   \,~\cite{35} \\ \hline
Complementary Colors      &   \,~\cite{15} & Global Texture                &   \,~\cite{8}  \\ \hline
Subordinate Colors        &   \,~\cite{15} & SDE &   \,~\cite{21} \\ \hline
Cooperate Colors          &  \, ~\cite{15} & Hue count                     &   \,~\cite{21} \\ \hline
Complexity feature        &   \,~\cite{15} & Depth of field                &   \,~\cite{7}  \\ \hline
\end{tabular}
\end{table}

\subsection{User-specific aesthetic distribution generation}
\label{usad_sec}

After acquiring the primary personalized ranking, we begin to focus on the user-specific aesthetic ranking that combine with both aesthetic rules and user's preference. Being aware of that primary personalized ranking is learned latently from user-specific images and hasn't dug deeply on why the users show preference in certain kind of images, we propose a well-designed style-specific aesthetic attribute classifier. Specifically, we investigate the high quality images both on professional photographic websites and the social network and find that visually-pleasing images always share certain aesthetic rules. Inspired by the task of multi-label classification, we proceed to generate an aesthetic distribution that integrate dozens of aesthetic attributes for all input images in the primary personalized ranking. Then, by concatenating distribution of each image in the ${R_{UPPR}}$, a final aesthetic distribution of the user is released. The aesthetic attributes we adopt are listed in Table ~\ref{aesthetic_tab}.

With the feature of elaborately designed aesthetic attribute generated in the aforementioned extracting stage, each input image with several  labels is transferred to our pre-trained one vs all classifier to generate a style-specific aesthetic distribution. The user-specific aesthetic distribution ${ {D_{USAR}} }$ is calculated by concatenating all the single style-specific distribution in Equation~\ref{eq5}.

\begin{equation}
\label{eq5}
{D_{USAR}} = \sum\limits_{i = 1}^C {\frac{{{r_C} - i + 1}}{{{r_C}\sum C }}{a_i}}
\end{equation}

In the equation above, ${r_C}$ is the ranking of \textsl{C-th} image's distribution. And ${a_i}$ represents aesthetic score of \textsl{i-th} set.

Since the user-specific aesthetic distribution is established from  primary ranking and multi-aesthetic attribute, there is no need to apply any internal strategies for capturing user's preference. Moreover, due to the learning of customized hyperplane, our framework avoids the instability suffered from the direct learning of user-specific images and enables an insightful and effective ranking.
 We adopt strategy shown below to generate user-specific aesthetic ranking:

\begin{equation}
S  =  \frac{{({D_{USAR}} - \overline {{D_{USAR}}} )\sum\limits_{i = 1}^n {({D_{test}} - \overline {{D_{test}}} )} }}{{\sqrt {{{({D_{USAR}} - \overline {{D_{USAR}}} )}^2}} \sqrt {\sum\limits_{i = 1}^n {{{({D_{test}} - \overline {{D_{test}}} )}^2}} } }}
\end{equation}

where ${ {D_{test}} }$ indicates the distribution of testing image, ${S}$ represents the relative score between ${D_{USAR}}$ and ${ {D_{test}} }$.

\begin{figure}[!t] \begin{centering}
  \centering
  \includegraphics[scale=.35]{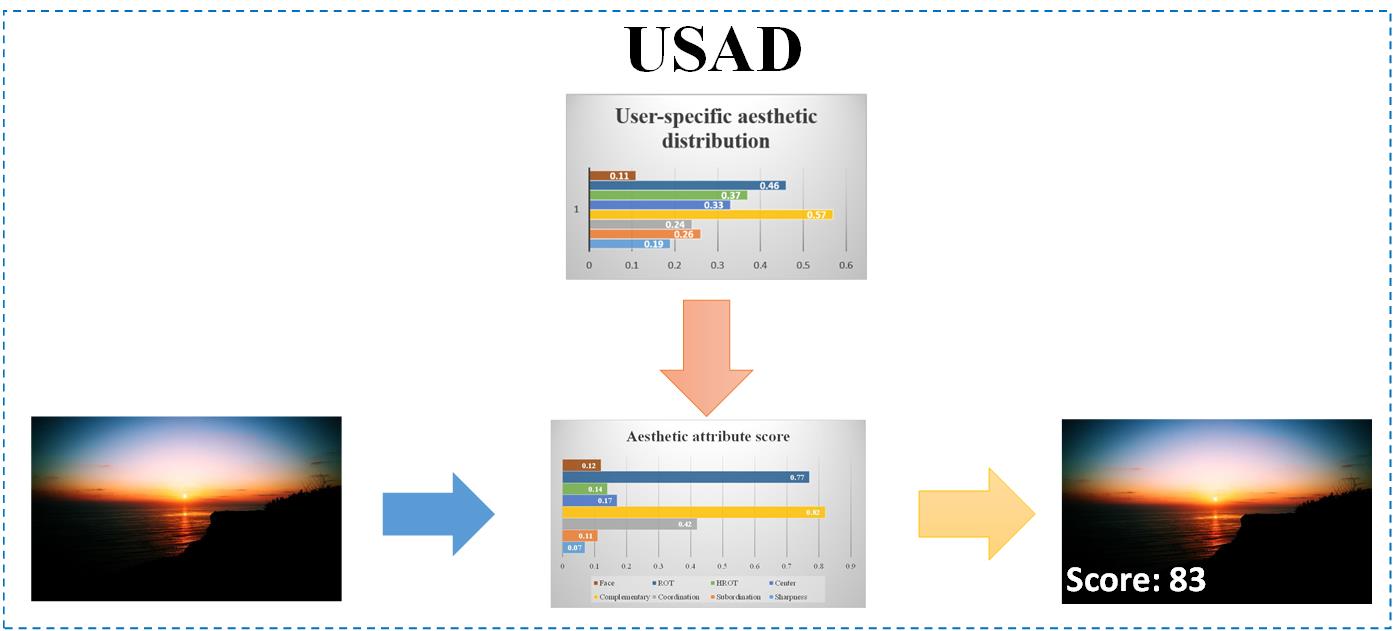}
  \centering
  \caption{The workflow during testing stage. For a given testing image, we send it to style-specific classifier to generate a testing aesthetic distribution ${D_{test}}$. Then, a score is deduced by calculating the correlation between testing distribution ${D_{test}}$ and user-specific aesthetic distribution ${D_{USAR}}$.}
  \label{test_stage}
  \end{centering}
\end{figure}

\subsection{Testing stage}

 After the user-specific aesthetic ranking (USAR) framework is learned during the training stage, we proceed to test our model as show in Fig.~\ref{test_stage}. Given a set of images to be tested, we conduct the following two tasks in parallel: 1) for user-specific images, we use our unique model to generate the relative rankings with real value scores. Specifically, for a given testing image, we send it to style-specific classifier to generate testing distribution ${D_{test}}$. Then, a real-valued score is deduced by calculating the correlation between ${D_{test}}$ and ${D_{USAR}}$. 2) we also push the images to the respondents, who are required to view and rate the same testing images as depicted in task 1. We use these respondent-generated scores as the ground truth in our experiment. Finally, we compare our prediction results with the ground truth to verify the effectiveness of our proposed model.

\section{Experiments}

In this section, we introduce how we conduct the experiment on two large scale public datasets and the comparsion with other state-of-art methods.

\subsection{Dataset}

The AVA dataset is a new large-scale dataset for conducting aesthetic visual analysis, which was collected by Murray et al.~\cite{29}. It contains over 250,000 images along with a rich variety of meta-data including a large number of aesthetic scores for each image collected during the digital photography contest on dpchallenge.com~\cite{30}. Each image on the website is commented and scored by users, commenter, participants and non-participants and the average score is provided. We treat the image with an average score of  5 plus as high quality and 5 minor as low quality as detailed in~\cite{28}. For the purpose of minimizing, we randomly shuffle the data ten times and assign half of them to training set and the rest to the testing set.

The FLICKR-AES dataset is constructed by 40000 photos with their aesthetic ratings marked by AMT,
which is downloaded from Flickr~\cite{39}. The aesthetic ratings range from the lowest 1 to the highest 5 to demonstrate the different aesthetic level. Each image in dataset is evaluated by five different AMT workers participated in annotation work of FLICKR-AES.

\subsection{Experimental setting and Method comparison}

In this section, we first detail the recruitment of the users. To validate the effectiveness of our user-specific aesthetic ranking learned with user's preference and ${D_{USAR}}$, we then compare the results with several state-of-the-art personalized aesthetic assessment method JR\_RSVR~\cite{38}, FPMF~\cite{35} and PAM~\cite{32}.

\subsubsection{User recruitment}

When choosing experiment volunteers, we take into
consideration the difference in age and occupation carefully. Since
the content of favorite images varies from the people with
different status, we choose volunteers aged from 23 to
45 (including college students, graduate students, teachers and
cleaners, etc.). In our experiment, the users are seated comfortably
in front of a computer. Our system will push images automatically for them to select. Once the user-specific images are
chosen, they are then sent to USAR\_PPR to learn customized
ranking function and a personalized ranking will be generated. Subsequently, interaction stage will be applied to
refine the primary personalized ranking with user's interaction.

\subsubsection{Comparison metric}

For the comparison with PAM and FPMF, we conduct experiments with the same experimental setting on FLIKR-AES dataset. The averaged results as well as the standard deviation are reported. To evaluate the performance on  AVA image dataset, we compare the result of ~\cite{33}, ~\cite{35} in terms of ranking correlation measured by Spearman's $\rho$ between the estimated aesthetics scores and the ground-truth scores. ${r_k}$ represents the k\textsl{-th} rank sorted by algorithm with the score ${S_k}$ and $\widehat {{r_k}}$ indicates the rank ordered by user with the score $\widehat {{S_k}}$. Subsequently, ${d_k} = {r_k} - \widehat {{r_k}}$ is then substituted by with $\rho  = 1 - \frac{{6\sum {d_k^2} }}{{{n^3} - n}}$ measuring the discrepancy between two ranks. $n$ represents the number of the images. The coefficient $\rho$ lies in range within $[- 1,1]$.


\begin{figure}[!t] \begin{centering}
  \centering
  \includegraphics[scale=.35]{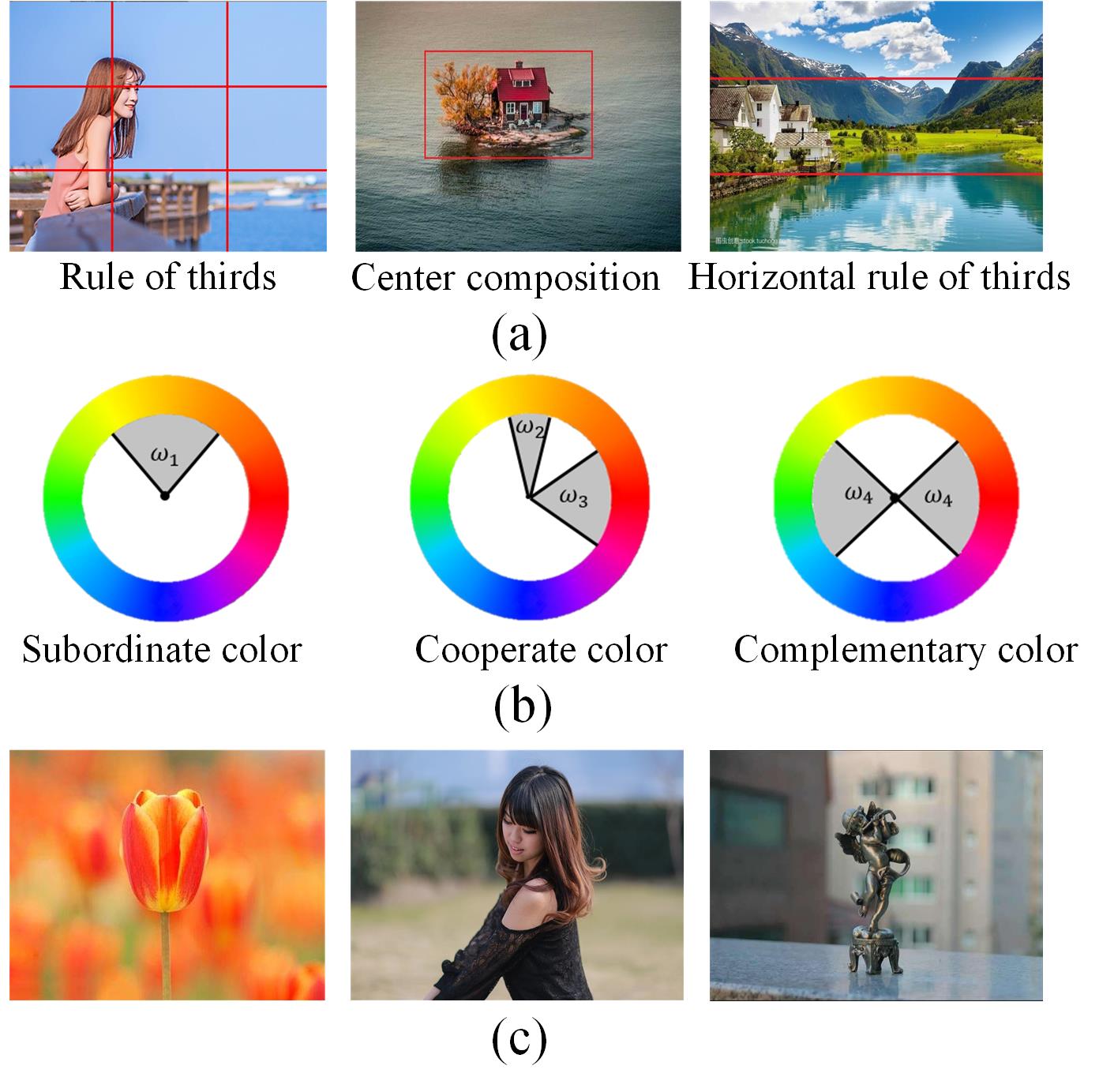}
  \centering
  \caption{Part of the metric images containing aesthetic attribute for users to refer to while annotating their personal ground-truth.}
  \label{d_metric}
  \end{centering}
\end{figure}

\subsubsection{USAR with different settings}

To thoroughly study and compare the various user-specific aesthetic ranking methods, we implement three methods and compare them with each other:

\begin{table}[]
\caption{Performance of our algorithm compared to other methods on FLICKR-AES Dataset}
\label{average_deviation}
\begin{tabular}{|l|c|c|}
\hline
                            & 10 images     & 100 images   \\ \hline
FPMF (only attribute)        & -0.003 $\pm$ 0.004 & 0.002 $\pm$ 0.003 \\
 \hline
FPMF (only content)          & -0.002 $\pm$ 0.002 & 0.002 $\pm$ 0.01  \\
\hline
FPMF (content and attribute) & -0.001 $\pm$ 0.003 & 0.01 $\pm$ 0.007  \\
\hline
PAM (only attribute)         & 0.004 $\pm$ 0.003 & 0.025 $\pm$ 0.013 \\
\hline
PAM (only content)           & 0.001 $\pm$ 0.004  & 0.021 $\pm$ 0.017 \\
\hline
PAM (content and attribute)  & 0.006 $\pm$ 0.003  & \textbf{0.039 $\pm$ 0.012} \\
\hline
USAR\_PPR                    & 0.003 $\pm$ 0.002  & 0.026 $\pm$ 0.007 \\
\hline
USAR\_PAD                    & 0.002 $\pm$ 0.003  & 0.019 $\pm$ 0.003 \\
\hline
\textbf{USAR\_PPR\&PAD}       &  \textbf{0.007 $\pm$ 0.004} & 0.034 $\pm$ 0.015 \\ \hline
\end{tabular}
\end{table}

${\rm{USAR\_PPR}}$ :  In user-specific aesthetic model, we first generate the ${R_{PPR}}$ by learned customized ranking function $f(\cdot)$. Generally, we acquire the user's preference by latently import preferred image to the system and denote the model as ${\rm{USAR\_PPR}}$ for short.

 ${\rm{USAR\_PAD}}$: For comparison with  ${\rm{USAR\_PPR}}$, we choose to directly use user-specific images as input to generate a concatenated aesthetic attribute distribution. For each testing image ${I_t}$, we send it directly to style-specific classifier and obtain a ${D_{test}}$. We generate the relative ranking by simply calculating its correlation and normalize it into certain range. The model above is marked by ${\rm{USAR\_PAD}}$ for short.

${\rm{USAR\_PPR\& PAD}}$: The objective of our proposed framework is to interactively generate a reliable user-specific aesthetic distribution. To achieve this, we organically concatenate the ${\rm{USAR\_PPR}}$ and ${\rm{USAR\_PAD}}$ with user's interaction. Specifically, given a set of images ${U_{test}} = ({I_{t1}},{I_{t2}},...,{I_{ti}})$ to be assessed, we first randomly select several images from the $\Gamma$ and present them to the users. By choosing several user-specific images, a primary personalized ranking ${R_{PPR}}$ is enforced by learned customized hyper plane. Considering the instability caused by a small number of samples, we apply the interactive strategy detailed in \textsl{Primary personalized ranking generation } and ask user to rerank the ${R_{PPR}}$ to acquire a new ${R_{PPR}}$ denoted by ${R_{UPPR}}$. Then we ask our respondents to view all the images on the testing dataset and rank the testing images in accordance with their preference. This is treated as the ground truth for our personalized ranking system. Meanwhile, the same testing images are sent to our proposed framework to generate a user-specific ranking ${R_{USAR}}$.

The comparison described above only involve the vertical comparison with different experimental settings. To validate the robustness of our proposed user-specific aesthetic distribution ${D_{USAR}}$, we implement the experiments 50 times on measuring the stability of the ${D_{USAR}}$ and present the correlation between ${D_{USAR}}$ and ground truth distribution ${D_{GTD}}$.

As for the ground-truth, we acquire the ground-truth ranking and distribution by inviting 20 users to participate in our test. All the users are required to rank the testing image in accordance with their preference. For the generation of the benchmark distribution, we ask the users to annotate certain aesthetic label with the benchmark metric shown in Fig.~\ref{d_metric}. For each user, we collect 50 sets of benchmark ranking and distribution marked by ${R_{GTR}}$ and ${D_{GTD}}$ and use it for measuring the performance with the different experimental settings.

\subsection{Result and discussion on FLICKR-AES dataset}

In this section,we compare our results with those of FPMF and PAM to verify the effectiveness of the proposed algorithm.
The average score as well as the standard deviation obtained on FLICKR-AES dataset with the number of 10 and 100 images is reported in Table~\ref{average_deviation}. Obviously, our three USAR models (last row in Table~\ref{average_deviation}) outperform the FPMF model. We then focus on the comparison between PAM and USAR. The data in second column represents the metric while selecting 10 images, which is identical to selecting 10 user-specific images. As shown in Table~\ref{average_deviation}, ${\rm{USAR\_PPR\& PAD}}$ outperform the PAM(content and attribute) while ${m =}$ 10. Even compared with the results of PAM when ${m =}$ 100, our ${\rm{USAR\_PPR\& PAD}}$ still demonstrates a competitive performance with the average improvement of 0.034, which validates the design of our user-specific aesthetic ranking model. The reason is that our model fully leverages the common understanding of multiple aesthetic attributes and user's preference, and well focuses on the interaction with ${R_{UPPR}}$ from the specific user.

We deem that excellent performance is achieved by two aspects: 1) ${\rm{USAR\_PPR}}$ initially learns the user's preference by unique retrieval result that are similar to user-specific images. Then the user's interaction enables a robust preference localization on stabilizing user's preference. Concretely speaking, distinguished from PAM~\cite{32}, our framework uses the input images selected by users to acquire the primary ranking firstly (USAR\_PPR part). After that, user interaction is involved to reorder the images in the primary ranking. This fine-tuned ranking is able to fit the very user's preference well. Finally, the well-tuned user-specific primary ranking is sent to the USAR\_PAD to generate stable aesthetic distribution. In summary, our method is obviously different from PAM by introducing interactive fine-tuning stage. This achieves an obvious performance enhancement meanwhile
keeps a relatively small amount of interaction. 2) The deployment of multiple aesthetic attributes enhance the performance initially deduced from ${R_{UPPR}}$ and enforce the ideal style-specific preference of the user.

We also calculate the ranking correlation introduced above to
measure the consistency between the prediction and ground-truth
data. Compared with the $\rho$ of 0.514 in ~\cite{32}, our mean ranking correlation over all the users is 0.518, which outperforms the
former.

\begin{figure*}[!t] \begin{centering}
  \centering
  \includegraphics[scale=.55]{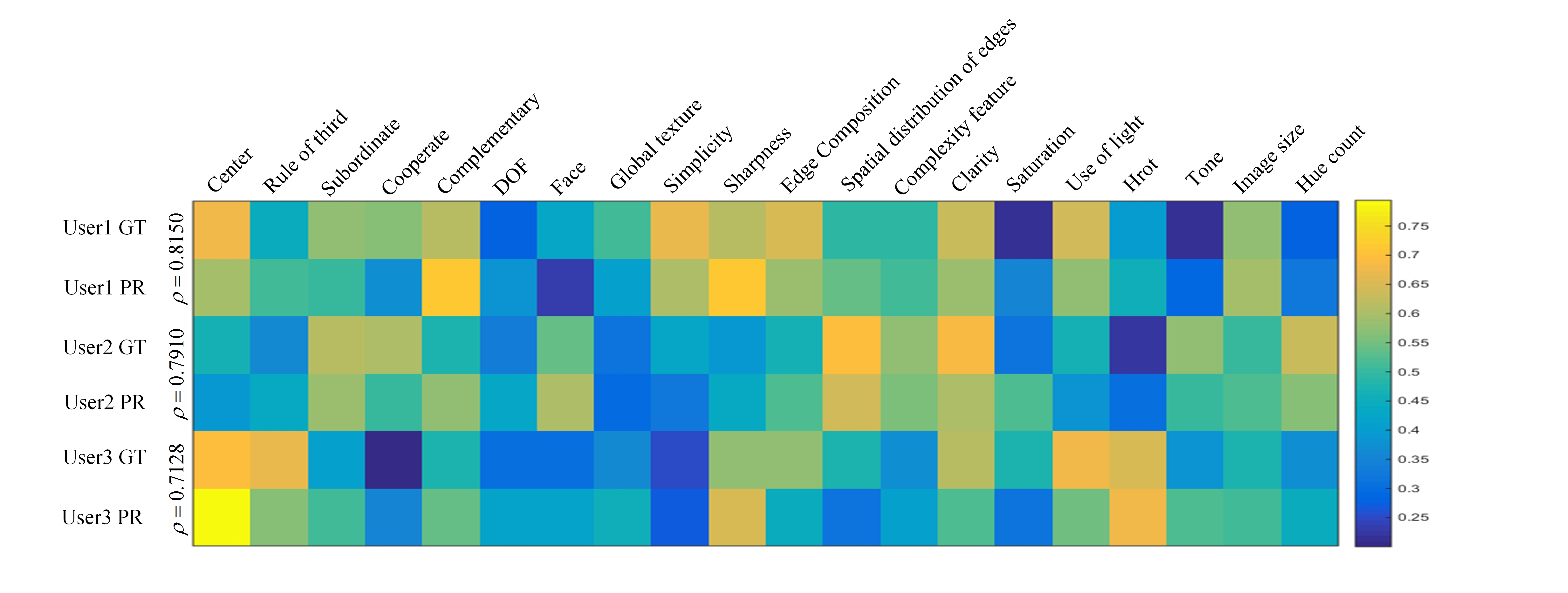}
  \centering
  \caption{The correlation between ground-truth aesthetic distribution and prediction aesthetic distribution of users. }
  \label{test_distribution}
  \end{centering}
\end{figure*}

\begin{table}[]
\centering
\caption{Performance of our algorithm compared to other methods on AVA Dataset. \textbf{USAR\_PPR\&PAD(5-5)} represents the setting when $m=$ 5, $k=$ 5}
\label{result_ava}
\begin{tabular}{|l|c|c|}
\hline
Method                  & $\rho$          & ACC(\%)        \\ \hline
Murray et al.           & -               & 68.00          \\ \hline
SPP                     & -               & 72.85          \\ \hline
RDCNN semantic          & -               & 75.42          \\ \hline
DMA\_AlexNet\_FT        & -               & 75.41          \\ \hline
JR\_RSVR                & 0.52            & -              \\ \hline
JR\_RSVM                & 0.30            & -              \\ \hline
AlexNet\_FT\_Conf       & 0.4807          & 71.52          \\ \hline
Reg+Rank+Att            & 0.5445          & 75.48          \\ \hline
Reg+Rank+Cont           & 0.5412          & 73.37          \\ \hline
Reg+Rank+Att+Cont       & 0.5581          & 77.33          \\ \hline
USAR\_PPR               & 0.6002          & 72.41          \\ \hline
USAR\_PAD               & 0.5446          & 77.69         \\ \hline
\textbf{USAR\_PPR\&PAD} & \textbf{0.5687} & \textbf{77.98} \\
\hline
\textbf{USAR\_PPR\&PAD(5-5)} & \textbf{0.5776} & \textbf{78.05}\\ \hline
\end{tabular}
\end{table}

\subsection{Results and discussion on AVA dataset}

The ranking correlation as well as accuracy on classification obtained on AVA Dataset are summarized in Table ~\ref{result_ava}. We first compare the proposed USAR model with the AlexNet\_FT\_Conf~\cite{33} and JR\_RSVM. The comparison indicates that our framework significantly outperforms the AlexNet\_FT\_Conf with a margin at 0.088 at least, regardless of the different settings of the USAR. We then check the comparison results with Reg+Rank+Att+Cont~\cite{33}, JR\_RSVR~\cite{35},  and USAR. The bottom seven rows in Table ~\ref{result_ava} show that proposed USAR\_PPR\&PAD with the Primary personalized ranking refined by user's interaction and Personalized aesthetic distribution yields the best performance (0.5687). Of the three models, the accuracy of USAR\_PPR (0.6002,72.41\%) is slightly weaker compared with that of Reg+Rank+Att (0.5445,75.48\%), while ranking correlation $\rho$(0.6002) is much stronger than that of the latter(0.5445). Aferwards, we focus on the results of USAR\_PAD. It is obvious that USAR\_PAD outperforms Reg+Rank+Cont with the margin on $\rho$ and ACC(\%) at 0.034 and 2.3. For the comparison with JR\_RSVR and Reg+Rank+Att+Cont, {USAR\_PPR\&PAD} shows the relative better performance both on ranking correlation and accuracy at 0.5687 and 77.98.
Despite USAR\_PPR\&PAD(5-5) yields the best performance with both ranking correlation(0.5776) and accuracy(78.05), it suffers from the poor experience induced by excessive interaction time and might cause aesthetic weariness.

Since the coefficient $\rho$ is calculated indirectly from the ${D_{USAR}}$, it is indispensable to evaluate the performance of ${D_{USAR}}$. We report the comparison between predicted distribution and the ${D_{GTD}}$ as shown in Fig.~\ref{test_distribution}. We list 3 of the 20 users and show the $\rho$ between ground-truth distribution(GT) and personalized distribution(PR) respectively. The comparison results show the high correlation with the maximum $\rho$ at 0.8150, which validates the effectiveness of the ${D_{USAR}}$.

We further explore whether or not the proposed ${\rm{USAR\_PPR\& PAD}}$ is capable of stabilizing user's preference with the limited choice of $m$ and $k$. To achieve this, we test the 20 users and calculate the ranking correlation $\rho $ between the ground-truth ranking and each individual's ranking with the varying parameter of $m$ and $k$ in Table~\ref{result_mk}.

\begin{table}[]
\centering
\caption{Ranking correlation measured by \textbf{$\rho$} with different $k$ and $m$}
\label{result_mk}
\begin{tabular}{|c|p{1.5cm}<{\centering}|p{1.5cm}<{\centering}|p{1.5cm}<{\centering}|}
\hline
                     & \multicolumn{3}{c|}{The number of user-specific images ($m$)} \\ \hline
Interaction times ($k$) & 5                       & 10             & 15             \\ \hline
1                    & 0.4869                  & 0.4823         & 0.4431         \\ \hline
2                    & 0.5341                  & 0.5389         & 0.5146         \\ \hline
3                    & 0.5687        & 0.5684         & 0.5427         \\ \hline
4                    & 0.5694                  & 0.5552         & 0.5345         \\ \hline
5                    & 0.5776                  & 0.5761         & 0.5644         \\ \hline
\end{tabular}
\end{table}

 Table~\ref{result_mk} yields the following observations. The ${\rm{USAR\_PPR\& PAD}}$ achieves the best performance with the $\rho$ at 0.5687 when $m = 5, k =3$, followed by 0.5776 with $m = 5, k = 5$. When compared with results on $m=10$ and $m=15$, the performance on $m=5$ outperforms the majority counterparts, where the maximum margin has reached
0.0438($m = 5,k = 1 $ and $m = 15,k = 1$). Even by increasing the interaction times $k$, the performance on $m=10$ and $m=15$ are still slightly weaker than that of $m = 5$. We declare this is due to two aspects:

 1) The inclination on selecting images while $m$ is relatively small. During the interaction stage, we observe our users and find that they tend to select the favorite images while $m$ is set relatively small, e.g. , $m=5$. The majority of five chosen images are of high quality (at least in user's opinion) or middle quality. This enables a robust learning to generate a reliable primary personalized ranking and the same to the user-specific aesthetic distribution. In contrast with the former, when $m = 10$ and more, users are inclined to pay more attention to the first several images. Specifically, the images ranked higher present the user's preference a bit more while that ranked lower present relatively less and mix the noise inevitably.

 2) The influence of different $k$. Throughout the whole result, the coefficient $\rho$ is positively related with trend of $k$. However, it begins to decrease while users choose 10 and 15 user-specific images with $k=4$. We believe that the sudden decrease suffers from over interaction with ${R_{PPR}}$. For one thing, the excessive interaction might cause the aesthetic fatigue. For another, it is impractical to conduct experiment with too much involved interaction.

Based on the aforementioned two reasons, we set $m = 5,k = 3$ and compare our results  with those of others on  AVA dataset as detailed above.


\section{Conclusion and future work}

We propose a novel and user-specific aesthetic ranking model which first combines the results of deep neural network with individual preference to explore a new overlap between the subjective feeling of the users and the aesthetic abstract. In our proposed ranking framework, we construct a reliable ranking framework consisting of Primary personalized ranking, Interaction stage and User-specific aesthetic distribution. Experimental results on two large datasets demonstrate the effectiveness and efficiency of our approach in two aspects: 1) Our framework obtains an excellent performance on predicting user's preference, whose results are closer to the personal intention. 2) Our framework enables a robust user-specific aesthetic distribution on user's preference and achieves relative high correlation when compared with previous work. Although we have achieved excellent performance by using Alexnet, which has a simpler structure to extract image features, it still takes a while for user to interact while we proceed to fine-tune the network again. In the future work, it is imperative to speed up the refinement so that consuming time of interaction stage could be reduced. Besides, the extracted aesthetic features in style-specific classifier are still hand-crafted features. In the future work, we manage to deploy stronger feature representation by adopting more powerful network structure, setting unique convolutional kernel and combining both hand-crafted and deep aesthetic feature in an attempt to achieve a more accurate user's personalized aesthetic preferences distribution.

\begin{acks}

The authors would like to thank Dr. Peng Pai from Youtu-Lab, Tecent Technology (Shanghai) Co.,Ltd from for his constructive suggestions to improve this work.

The authors would also like to thank the anonymous referees for
  their valuable comments and helpful suggestions. This work is
  supported by the \grantsponsor{GS501100001809}{National Natural
    Science Foundation of China}{http://dx.doi.org/10.13039/501100001809} under Grant
  No.:~\grantnum{GS501100001809}{61772474, 61672469}. Pei Lv and Meng Wang contribute to this work equally.


\end{acks}


\bibliographystyle{ACM-Reference-Format}
\bibliography{sigproc}

\end{document}